\documentclass[11pt]{article}

\usepackage[utf8]{inputenc} % allow utf-8 input
\usepackage[T1]{fontenc}    % use 8-bit T1 fonts
\usepackage{hyperref}       % hyperlinks
\usepackage{url}            % simple URL typesetting
\usepackage{booktabs}       % professional-quality tables
\usepackage{amsfonts}       % blackboard math symbols
\usepackage{nicefrac}       % compact symbols for 1/2, etc.
\usepackage{microtype}      % microtypography
\usepackage{graphicx}
\usepackage{textgreek}
\usepackage{comment}
\usepackage{epstopdf}
\usepackage[caption=false]{subfig}
\usepackage[numbers,sort&compress]{natbib}
\usepackage{amsmath}
\usepackage{amssymb}
\usepackage[margin=1in]{geometry}

\graphicspath{ {./figures/} }

\bibpunct[, ]{[}{]}{,}{n}{,}{,}

\makeatletter
\def\NAT@def@citea{\def\@citea{\NAT@separator}}
\makeatother

\newenvironment{keywords}{\begin{quote}\textbf{Keywords: }}{\end{quote}}

\title{Designing a double deep reinforcement learning selection tool for resilient demand prediction}

\author{
Bilel Abderrahmane Benziane\textsuperscript{c}\thanks{CONTACT Bilel Abderrahmane Benziane Author. Email: babenziane@generixgroup.com}, Benoit Lardeux\textsuperscript{b}, Ayoub Mcharek\textsuperscript{c} and Maher Jridi\textsuperscript{a} \\
\textsuperscript{a}L@bISEN, Vision-AD, ISEN Yncréa Ouest, 33 Quatre Chemin du Champ de Manoeuvre, 44470, Carquefou, France \\
\textsuperscript{b}L@bISEN, Usine du Futur, ISEN Yncréa Ouest, 33 Quatre Chemin du Champ de Manoeuvre, 44470, Carquefou, France \\
\textsuperscript{c}DataLab, Generix Group, 8 Rue Simone IFF, 75012, Paris, France
}

\date{}

\begin{document}

\maketitle
\noindent\textbf{Published in} \textit{Journal of Industrial and Production Engineering}. DOI: \href{https://doi.org/10.1080/21681015.2025.2580997}{10.1080/21681015.2025.2580997}\\
\begin{abstract}
The use of artificial intelligence in supply chain forecasting has attracted many scientific studies for several decades. However, the process of selecting an appropriate forecasting solution becomes a daunting task. This complexity arises due to the distinct features inherent to each dataset. Research to tackle this issue has been performed since the eighties but recent development of demand forecasting has opened new perspectives. This research aims to enhance automatic forecasting model selection by proposing a novel architecture that acts as a double deep reinforcement learning agent, selecting automatically a forecasting model from the forecasting committee at the time of prediction. Moreover, a novel early-stopping approach based on average reward convergence has been introduced to expedite training time. To evaluate the model's performance, an empirical study was conducted utilizing grocery sales datasets and snack demands datasets. The experimental results demonstrate the robustness of the proposed approach when compared to state-of-the-art methods.
\end{abstract}


\section{Introduction}
Recent events such as the Covid-19 pandemic and ongoing regional conflicts between nations across several regions of the world have significantly altered the landscape of demand data \cite{10.1145/3456529.3456539,doi:10.1177/0022343312449033}. Traditional inventory management techniques are no longer sufficient to cope with these changes. In response, a plethora of computing solutions and software tailored for managing the flow of goods and the means to convey or stock them have emerged. These solutions aim to closely monitor the real-time status of company supply chains, making them indispensable for ensuring continuous and appropriate fulfillment of sales endpoints. The study presented in this paper has been performed to design new software solutions for an IT company dedicated to supply chain optimization. A key initial step for computing recommendations in supply chain optimization involves being able to anticipate customer demands for goods consumption. Based on this information and the many other supply chain parameters, the software can recommend required levels of stocks and plans for product transportation towards consumption zones. Demand prediction is therefore a necessary step for supply chain management, as outcomes of this step may impact several subsequent stages of the supply chains. Accurate predictions allow companies managing supply chains to cut down business costs generated by overstocking or missing sales opportunities due to underestimation of customer demand. \\

In the literature, many artificial intelligence forecasting architectures have been designed \cite{10.1007/978-3-030-28377-3_27,kiefer_artificial_2022,chaudhuri_hybrid_2022,abhishekh_new_2020,benziane_supply_2023,10912530,doi:10.36227/techrxiv.174910155.59321144/v1,doi:10.36227/techrxiv.174235483.35557269/v1,benziane:tel-05503809,hadjsalem:hal-05060572}.
Despite the profusion of forecasting methods, no single method can consistently perform equally well or poorly across all use cases. This variation arises due to the influence of numerous factors that can affect the quality of forecasts, particularly those related to data. Data reveals diverse features, such as high seasonality, substantial bias, noise, volatility, intermittence, and more, each type of data requiring a specific model for effective handling \cite{FU2019940,finco_applying_2022,ABOLGHASEMI2020106380,benziane_supply_2023,MURRAY201833}. Taking into consideration that data characteristics can change in real-time, this can substantially impact forecast quality. Numerous other data-related problems have been highlighted in \cite{GONZALEZPEREA201959} as the main reasons that may alter forecast quality.  This hypothesis has been verified by Angos et al. in their review of forecast papers from 2017 to 2021 where they have shown that the best-performing models change based on data characteristics, such as dimension and volume \cite{mediavilla_review_2022}. \\

To mitigate the impact of data-related problems, various methodologies have been proposed. In every case, authors prioritize methods that deal with specific problems based on the use case or the targeted business application. In this context, optimization methods are applied to assess and compare multiple configurations of the same architecture, so that the configuration with the smallest error is selected \cite{chaudhuri_hybrid_2022}, \cite{GONZALEZPEREA201959}. Additionally, models tailored to specific business needs are designed in the following articles \cite{GONZALEZPEREA201959}, \cite{8939050}, \cite{10.1007/978-3-030-15035-8_107}. It's important to note that each of these solutions is designed for a specific dataset or problem. Ensembling techniques, where multiple models are combined, are also widely used in the forecasting field \cite{9358156}, \cite{7178838}, \cite{yan_multi-step_2018}, \cite{JOSEPH2022108358}. They have been proven to be efficient whenever data sources of different types input predictions. However, if the selection of the appropriate solution approach is not frequently reviewed, the errors generated by the model would be inclined to increase over time. \\

To address the need for forecasting solutions to be agile and adaptable to changing data dynamics, this study aims at developing an optimized double deep convolution Q-learning algorithm with average reward improvement based early stopping approach. This algorithm will be able of continuously selecting a forecasting model among multiple neural network-based models. The main contributions of this work are described as follows

\begin{enumerate}
    \item Proposition of a double deep reinforcement learning for forecasting model selection: Our primary contribution is the introduction of a double deep reinforcement learning (DDRL) architecture designed to dynamically select forecasting models at any given time. This model runs over the historical demand history and the forecasted values of the entire forecasting committee as its state, which undergoes processing through a convolutional neural network (CNN), recurrent neural networks (RNN), and a feedforward neural network (FFNN) for spatial and temporal feature extraction, enabling informed decision-making.

    \item Introducing an average reward-based early stopping: Our second contribution involves introducing a novel technique for early stopping based on the improvement rate of the average reward. This method aims to reduce training time significantly.

    \item Assessment with an empirical study: Lastly, we present an empirical study that rigorously evaluates and compares the performance of the proposed approach against both the forecasting committee and four automatic model selection methods. This analysis is conducted on two distinct datasets: a publicly available grocery sales dataset and a snacks demand dataset provided by a client in France. The results obtained from this study will showcase the effectiveness and robustness of the proposed solution across various data settings.
\end{enumerate}

The rest of this paper is split into five sections: Section 2 is the Literature Review, Section 3 explains the Methodology (Explanation of the empirical study methodology), Section 4 discusses the Experimental Setup (Details of the experimental setup), Section 5 presents the Results and Discussion from experiments, and Section 6 concludes this work and includes proposals for following research directions. This structure provides a consistent flow of the research, allowing readers to follow the development and validation of the proposed forecasting model.

\section{Literature review}

Forecasting based on analysis of history is a problem that has been tackled for several decades. In many industries, guessing customer behavior is crucial for ensuring appropriate business returns, such as flight inventory management, distribution, and supply chain optimization. In recent years, many different methodologies focusing on the artificial intelligence paradigm have been proposed for general forecasting purposes. This literature review aims at summarizing the primarily used forecasting models, the methodologies selected to harness these models, and various approaches for the selection of forecasting models. Ultimately, our research endeavors to address the identified research gaps from recent articles by proposing a solution combining different neural network architectures selected according to the context by reinforcement learning.

\subsection{Models}

In the past, classical forecasting methods such as Exponential Smoothing (ETS) \cite{brown1956exponential} and Autoregressive Integrated Moving Average (ARIMA) \cite{box1970time} have been the dominant approach in the forecasting domain, owing to two significant benefits. Firstly, these methods have relatively low requirements for training data; this was a crucial factor during that period, as most businesses collected data at lower frequencies such as monthly or yearly. Secondly, classical methods have minimal computational requirements, which was especially important given the limited computational power of computers at that time. These factors made classical methods an attractive option, while the more complex artificial intelligence methods were largely ignored due to their lack of suitability under these constraints. \\

In recent years, computer performance and data availability were no longer an issue. Back in 2011, it was shown that neural networks and computational intelligence methods can perform competitively with regards to the established classical methods in time series prediction \cite{crone_advances_2011}. This led to many researchers developing and implementing more sophisticated forecasting models, particularly those mentioned below.

\subsubsection{Artificial Neural Networks (ANNs)}

The strength of ANNs lies in their capability to effectively capture non-linearity and intricate patterns. Perea et al. employed ANNs to forecast water demand for irrigation using a short dataset \cite{GONZALEZPEREA201959}. Guven et al. applied ANNs for predicting demand in the apparel industry through two distinct approaches, whereas ANN outperformed support vector machines (SVM) in both experimental setups \cite{guven_demand_2020}.

\subsubsection{Recurrent Neural Networks (RNNs)}

Recurrent Neural Networks (RNNs) excel in the detection of sequential data patterns \cite{LSTM_paper}. The benefits of this approach have been shown in several recent papers: Hodžić et al. applied Long Short-Term Memory (LSTM) networks for the forecasting of warehouse demand \cite{8939050}. Kic et al. utilized LSTM to predict the demand for medical equipment in Turkey during the COVID-19 pandemic \cite{koc_forecasting_2022}. Chandriah et al. applied LSTM with a modified Adam optimizer to achieve commendable results in spare parts demand forecasting \cite{chandriah_rnn_2021}. Bandara ran LSTM in conjunction with clustering to overcome traditional forecasting methods such as auto-regressive integrated moving average (ARIMA) \cite{bandara_forecasting_2020}.

\subsubsection{Convolutional Neural Network CNNs}

Other neural network architectures have been tested and are proven to be efficient in some special cases. Convolutional Neural Networks (CNNs) for instance are widely applied in forecasting due to their ability to autonomously learn feature engineering through filter optimization. Khan, S designed Deep CNNs for electricity load forecasting in Australia, achieving significant performance in accommodating the challenging nature of the data, characterized by high volatility, non-stationary, and non-linear behavior of electricity load data. The CNN model outperformed LSTM in this context \cite{10.1007/978-3-030-15035-8_107}.

\subsubsection{Ensembling forecasting}

In previous paragraphs, we noticed that different kinds of network architectures perform well according to the context and the nature of data. Therefore approaches combining different architectures seem to be promising solution approaches whenever the forecaster needs to minimize errors given various data contexts. Ensemble modeling is a process in which multiple diverse base models are used to predict an outcome. The rationale behind using ensemble models is to mitigate the generalization error of predictions. \\

One strategy for ensemble modeling is boosting, in which models are sequentially combined. In this approach, each model's output serves as the input for the next model. For instance, Shafiul et al. proposed a combination of CNN and LSTM models. Initially, CNNs were employed to capture local trends in the load data pattern, and their outputs were subsequently fed into LSTM models to capture long-term dependencies within the load dataset. The final forecasting results were generated after a complete connection through dense layers \cite{9358156}. \\

Sainath et al. took a different approach, combining CNNs, LSTMs, and DNNs (Deep Neural Networks) into a unified architecture known as CLDNN. They assessed this architecture on a range of large vocabulary tasks, spanning from 200 to 2,000 hours, and found that CLDNN exhibited a 4-6 percent relative improvement in Word Error Rate (WER) over LSTM, which was the strongest among the three individual models \cite{7178838}. Ke Yan et al. improved electric power consumption short-term forecasting by applying a combination of CNN and LSTM models \cite{yan_multi-step_2018}. Additionally, Joseph et al. used boosting with CNN and bidirectional LSTM models to forecast the Store Item Demand Forecasting Challenge dataset from Kaggle \cite{JOSEPH2022108358}. \\

Another ensemble technique is stacking, where models run in parallel, and their results are fed into a meta-model that learns to generate the final forecast based on the inputs from different models. Yan et al. applied stacking to multiple configurations (varying the number of neural network layers) for electric energy consumption forecasting \cite{MOON2020109921}.

\subsection{Data problems related solutions}

Forecasting different datasets, meaning different time series each one having its own overall features (seasonality, patterns, etc...) may require the use of various models, primarily due to the distinct data challenges they present, such as intermittence and volatility, as highlighted by Abolghasemi et al. \cite{ABOLGHASEMI2020106380}. Determining a single model that consistently outperforms others is challenging, prompting the emergence of various research endeavors which aim at addressing this issue.

\subsubsection{Selecting best hyperparameters for one model - optimization}

Optimization techniques play a crucial role in identifying the best configuration for a given architectural model. Abbasimehr et al. designed a grid search to optimize LSTM and determine the best model configuration \cite{ABBASIMEHR2020106435}. Perea et al. developed a Genetic Algorithm (GA) optimizer for optimizing ANN in the context of water demand forecasting for irrigation, particularly on short datasets \cite{GONZALEZPEREA201959}. Chaudhuri et al. harnessed the Harris Hawk Optimization method to discover the optimal configuration of Extreme Learning Machines (ELMs) in the context of product demand forecasting applications \cite{chaudhuri_hybrid_2022}. While testing various models can indeed enhance performance, it is important to acknowledge that the optimal model selection can vary depending on the specific features of the data at hand.

\subsubsection{Selecting best model}

The selection of forecasting models traditionally relies on domain experts who have experimented over time with the best approaches according to the context. However, human interventions can be time-consuming and prone to errors, even if they are generally complex to replace by machines. Moreover, this approach may not be efficient for businesses with prior knowledge of the AI field. \\

Fu et al. introduced an approach involving exponentially weighted regret combinations. In this method, each model is assigned a weighted value, and forecast errors are computed. Decision regret for each model at time t is then calculated, followed by the update of the weighted vectors. This technique ensures that more accurate forecasting models contribute more significantly to the forecasting process \cite{FU2019940}. Villegas et al. applied a Support Vector Machine (SVM) for model selection, which resulted in more robust predictions with lower mean forecasting errors and biases compared to base forecasts \cite{VILLEGAS20181}. \\

Recognizing that data features can evolve over time, the optimal forecasting model may change as well. To address this issue, the use of reinforcement learning for model selection has been proposed. Dabbaghjamanesh et al. included Q Learning which selects between Feed-Forward Neural Networks (FFNN) and Recurrent Neural Networks (RNN) for load forecasting under different scenarios, leading to reduced forecasting errors \cite{9078831}. Al haj Hassan et al. applied reinforcement learning for weighted model selection whenever the final forecast is a combination of multiple weighted sums from a committee of forecasters. The weight percentages are computed based on each forecaster's past error compared to the total error of all forecasters \cite{al_hajj_hassan_reinforcement_2020}. \\

In this study, we propose reinforcement learning with a new architecture relying on convolutional, recurrent, and feedforward layers (CRFFNN) as a double deep reinforcement learning agent to better capture the spatial and temporal features of the environment state, ultimately reducing selection errors. To tackle the potentially time-intensive nature of training, we suggest an Average Reward Improvement Rate Based Early Stopping (ARIRBES) approach to reduce training time and costs. Through multiple comparative experiments, our proposed architecture achieves more accurate and robust results in studied cases from grocery sales, as this will be shown in the section of the empirical study. Before we detail further the proposed model and solution approach in the next section.

\section{Methodology}

In this section, the concepts of deep reinforcement learning are introduced. Then, the proposed architecture that is called CRFFNN-ARIRBES is explained in details as this is the core of this paper contribution.

\subsection{Concept of Double deep reinforcement learning}

Forecasting models, like all tools, come with their strengths and weaknesses, making it challenging to identify a universal approach that consistently produces excellent results for every cases. Typically, it falls upon forecasting specialists to make this selection, a process that can be time-consuming and prone to errors. Fortunately, there are various methods to automate this selection process, and one standout approach is reinforcement learning. Reinforcement learning historically performs well in adapting to noisy environments with constantly changing behavior. In this paper, we propose an architecture based on Q-learning, which is a foundational algorithm in the field of reinforcement learning. Q-learning has been designed to guide an agent's decision-making within a Markov decision process (MDP), with the ultimate goal of maximizing cumulative rewards during interactions with its environment. This groundbreaking algorithm, introduced by Christopher J. Watkins in his 1989 Ph.D. thesis, is widely acclaimed for its significance in advancing the field of reinforcement learning \cite{dql_article}. \\

\noindent The Q-learning equation (\ref{eq:equation_ql_1}) is a fundamental component of this algorithm and is used to update the Q-values, which represent the expected cumulative rewards an agent can achieve by taking a specific action in a particular state.
\begin{equation}
Q(s, a) = Q(s, a) + \alpha \cdot [r + \gamma \cdot \max(Q(s', a')) - Q(s, a)]
\label{eq:equation_ql_1}
\end{equation}
where :
\begin{enumerate}

    \item $\alpha$ (alpha) is the learning rate, controlling how much the Q-values are updated based on new information.
    \item $\gamma$ (gamma) is the discount factor, which determines the importance of future rewards. It is a value between 0 and 1.
    \item $s'$ is the new state after taking action $a$.
    \item $a'$ is the action selected in the new state $s'$.
    \item $Q(s,a)$ represents the Q-value of taking action $a$ in state $s$. Q-values are used to estimate the expected cumulative rewards when starting in state $s$, taking action $a$, and then following the optimal policy thereafter.
    \item $\max(Q(s',a'))$ represents the maximum Q-value over all possible actions $a'$ in the next state $s'$. It estimates the maximum expected cumulative reward the agent can achieve from the next state onward, assuming it follows the optimal policy.
\end{enumerate}

\noindent The Q-learning algorithm repeatedly applies this equation to update the Q-values for state-action pairs as the agent explores and learns from its interactions with the environment. Over time, the Q-values converge to their optimal values, and the agent can use these values to select the best actions at every state, ultimately leading to an optimal policy for maximizing cumulative rewards in the MDP. \\

The task of reinforcement learning in the platform involves selecting a forecast model based on the computed forecasts of all candidates. When dealing with an extended forecast horizon encompassing decimal numbers, the number of possible state-action combinations becomes overwhelmingly large, requiring significant computational power. As a result, this study focuses on a subset of reinforcement learning made by deep neural networks to approximate either the value function or policy. This concept is known as deep reinforcement learning (Deep RL). \\

\begin{figure*}[htbp]
    \centering
    \includegraphics[width=0.9\textwidth, angle=0, trim={0cm 2cm 0cm 0.5cm}, clip]{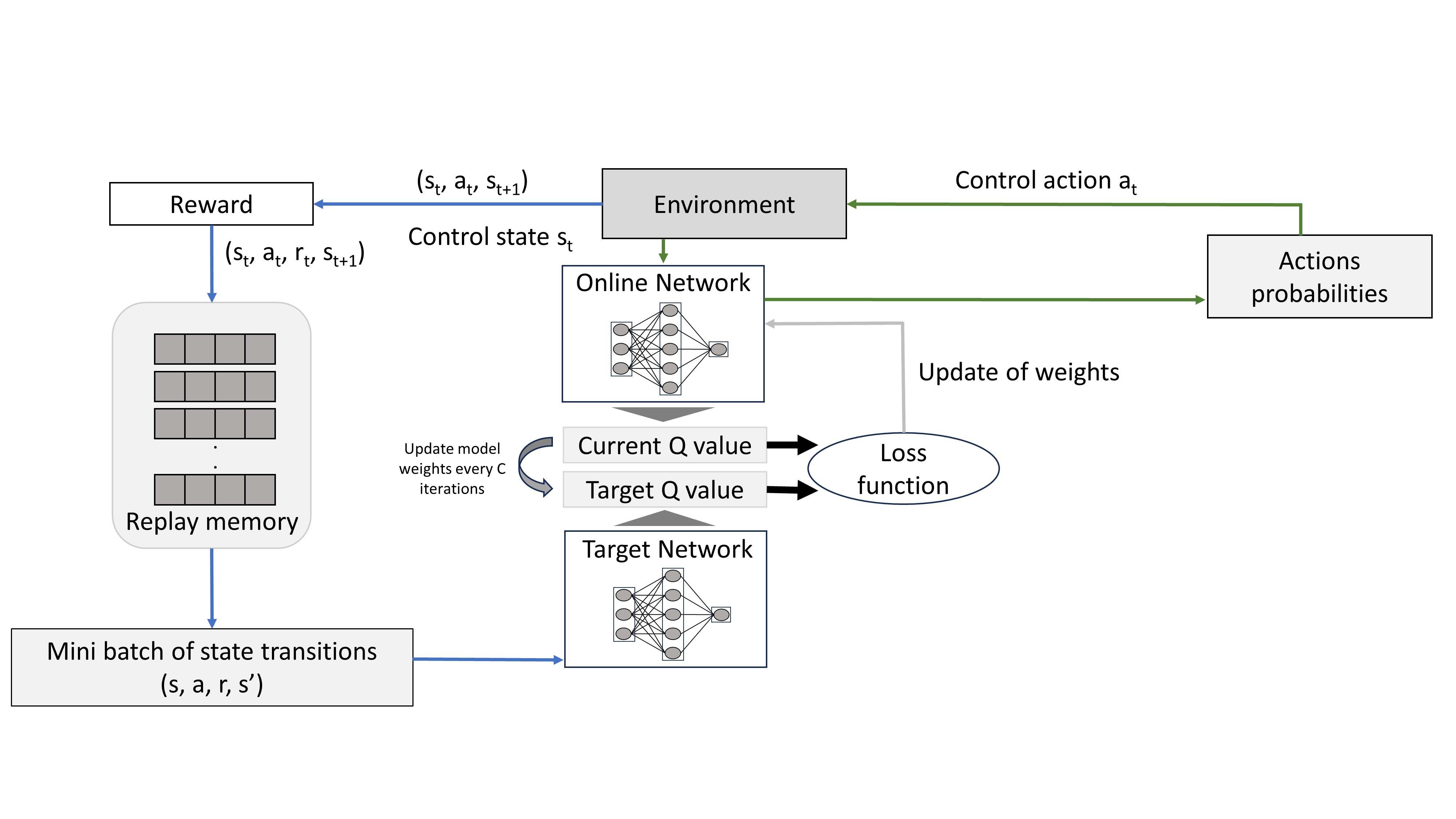}
    \caption{Double deep Q learning architecture.}
    \label{fig:methodology_9_ddql}
    \vspace{-0.1cm}
\end{figure*}

In Deep RL, Q-learning is a commonly used technique, where Q-values are estimated using a deep neural network, often referred to as a Q-network \cite{dql_article}. This Q-network takes the current state as input and produces Q-values for each available action as output. However, it's essential to note that Deep RL methods can sometimes lead to overestimation of Q-values, which can result in sub optimal or unstable training. To address this overestimation issue in Deep RL, we introduce Double Deep RL, a technique outlined in \cite{ddql_paper}. In Double Deep RL, as shown in Figure \ref{fig:methodology_9_ddql}, two separate Q-networks, denoted as the Online Q network and Target Q network are used to estimate Q-values. During the process of target Q-value estimation and action selection, these networks are alternately activated. By adopting this alternating approach with the two networks, Double Deep RL mitigates the tendency to overestimate Q-values, leading to more stable and improved performance of reinforcement learning agents.

In Double Deep Q Learning (DDQN), the process involves several key elements:
\begin{enumerate}
    \item \textbf{Q-Networks:} \begin{enumerate}
            \item \textbf{Online Q-Network:} This neural network estimates Q-values, which represent the expected cumulative rewards for taking various actions in the current state. It's updated during the training process.
            \item \textbf{Target Q-Network:} The target Q-network is used to calculate the expected future rewards but is updated less frequently. It serves as a more stable reference for the online Q-network  (every C iteration).     \end{enumerate}
    \item \textbf{Loss function:} The loss function quantifies the error between the Q-values predicted by the online Q-network and the Q-values estimated by the target Q-network. Minimizing this loss is the first objective during training.
    \item \textbf{Current state and future state:}
    \begin{enumerate}
        \item The "current state" \textbf{S} is the agent's current configuration within the environment. It is the demand of the previous 35 days concatenated with the output of the six forecasters committee \[
s = [d_{t-35}, d_{t-34}, \dots, d_{t-1}, f_{t1}, f_{t2}, \dots, f_{t6}]
\] Where, $\textbf{d}$ represents the demand per day, and $\textbf{f}$ represents the forecaster output.
        \item The "future state" $\textbf{S'}$ is the state the agent transitions into after taking a specific action in the current state. For instance, if the DDQL agent selected the second forecasting model, the future state would be as follows \[s'=
[d_{t-34}, d_{t-33}, \dots, d_{t-2},f_{t2}, f_{t1}, f_{t2}, \dots, f_{t6}]
\].
    \end{enumerate}
    \item \textbf{Reward:} The agent receives a numerical reward (\ref{eq:reward}) from the environment after taking an action in a given state. This reward quantifies the immediate benefit or penalty of the action. Including the exponential in the denominator means having a smaller reward when the error is big.
        \begin{align}\label{eq:reward}
           \text{reward} &= \frac{1}{e^{\text{mse}(Y_{\text{actual}}, Y_{\text{predicted}})}}
        \end{align}
    \item \textbf{Update Target Q-Network:} To mitigate overestimation of Q-values, the target Q-network is periodically updated, but not after every time step. This update involves copying the weights from the online Q-network to the target Q-network, maintaining a lagged version of the online network.
\end{enumerate}

The DDQN training process unfolds as follows:
\begin{enumerate}
    \item Initialization: Begin by initializing both the online and target Q-networks. Configure hyperparameters like the learning rate, and discount factor.
    \item Action Selection: At each time step, the agent selects an action according to an exploration strategy, typically epsilon-greedy. The chosen action is executed in the environment, and the agent observes the resulting reward and the subsequent state.
    \item Update Q-Values:
    Calculate the Q-value for the current state-action pair using the online Q-network.
    Estimate the Q-value for the future state using the target Q-network.
    Compute the target Q-value, which is the sum of the observed reward and the maximum Q-value for the future state.
    Calculate the loss as the discrepancy between the target Q-value and the current Q-value predicted by the online Q-network.
    \item Backpropagation: Update the weights of the online Q-network via backpropagation according to the computed loss.
    Periodic Target Network Update: Periodically update the target Q-network by copying the weights from the online Q-network. This maintains a consistent reference for Q-value estimation.
    Iterative Learning: Repeat steps 2-4 for multiple episodes or until the agent's policy converges or a predefined termination condition is met.
\end{enumerate}

\subsection{Convolutional Recurrent Feed-Forward Neural Network (CRFFNN)}

Instead of relying on a straightforward feedforward neural network to interpret the outputs of the forecasting committee and select the optimal model, this research proposes a more sophisticated approach. We advocate for a deep neural network architecture including convolutional, recurrent, and feedforward modules. This architecture is designed to enhance the modeling of sequential data by capturing effectively both spatial and temporal features. \\

The initial convolutional module extracts spatial features inherent in the dataset, enabling the model to discern complex spatial patterns. Following this, the recurrent module comes into play, demonstrating its efficacy in capturing temporal patterns within the data. By leveraging the recurrent layer, the model gains the capability to appreciate and exploit time-dependent relationships, further enhancing its predictive capacity. Lastly, the feedforward network serves to consolidate these extracted features and map them to the expected output shape, easing the generation of accurate forecasts. This architecture ensures that the model can handle the intricacies of sequential data, incorporating both spatial and temporal elements for improved performance. \\

Moreover, rather than only relying on the outputs of the forecasting committee, our approach integrates information from the past 35 observations as well. This inclusion allows the model to glean additional context from the environment, providing it with a more comprehensive understanding of the underlying data dynamics. By considering both historical observations and committee outputs, the model is better equipped to make informed decisions, leading to more robust and accurate forecasting outcomes. \\

\begin{figure*}[htbp]
    \centering
    \includegraphics[width=0.9\textwidth, angle=0, trim={2cm 2.5cm 2.5cm 0.5cm}, clip]{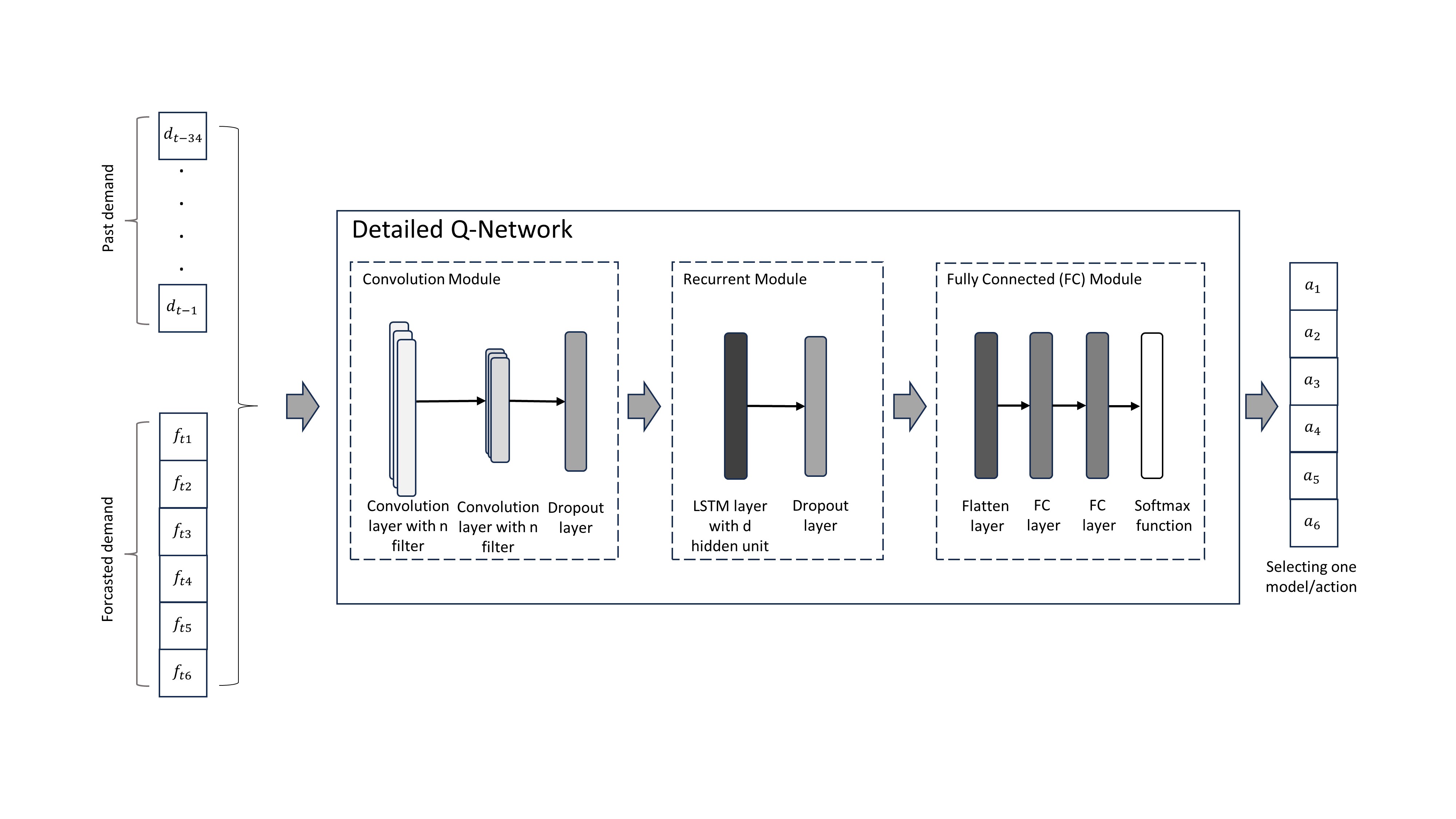}
    \caption{The suggested CRFFNN architecture.}
    \label{fig:methodology_ddql_agent}
    \vspace{-0.1cm}
\end{figure*}

As depicted in Figure \ref{fig:methodology_ddql_agent}, the methodology starts by integrating the past 35 observations
 \([d_{t-35}, \dots, d_{t-1}]\), thereby providing the model with key historical context. These observations are subsequently concatenated with the forecasts generated by the forecasters committee \([f_{t1}, \dots, d_{t6}]\), facilitating the amalgamation of historical data and ensemble predictions within the model input.

\subsection{Average Reward Improvement Rate Based Early Forecasting (ARIRBEF)}

\begin{figure*}[htbp]
    \centering
    \includegraphics[width=0.9\textwidth, angle=0, trim={2cm 1.8cm 0cm 2.3cm}, clip]{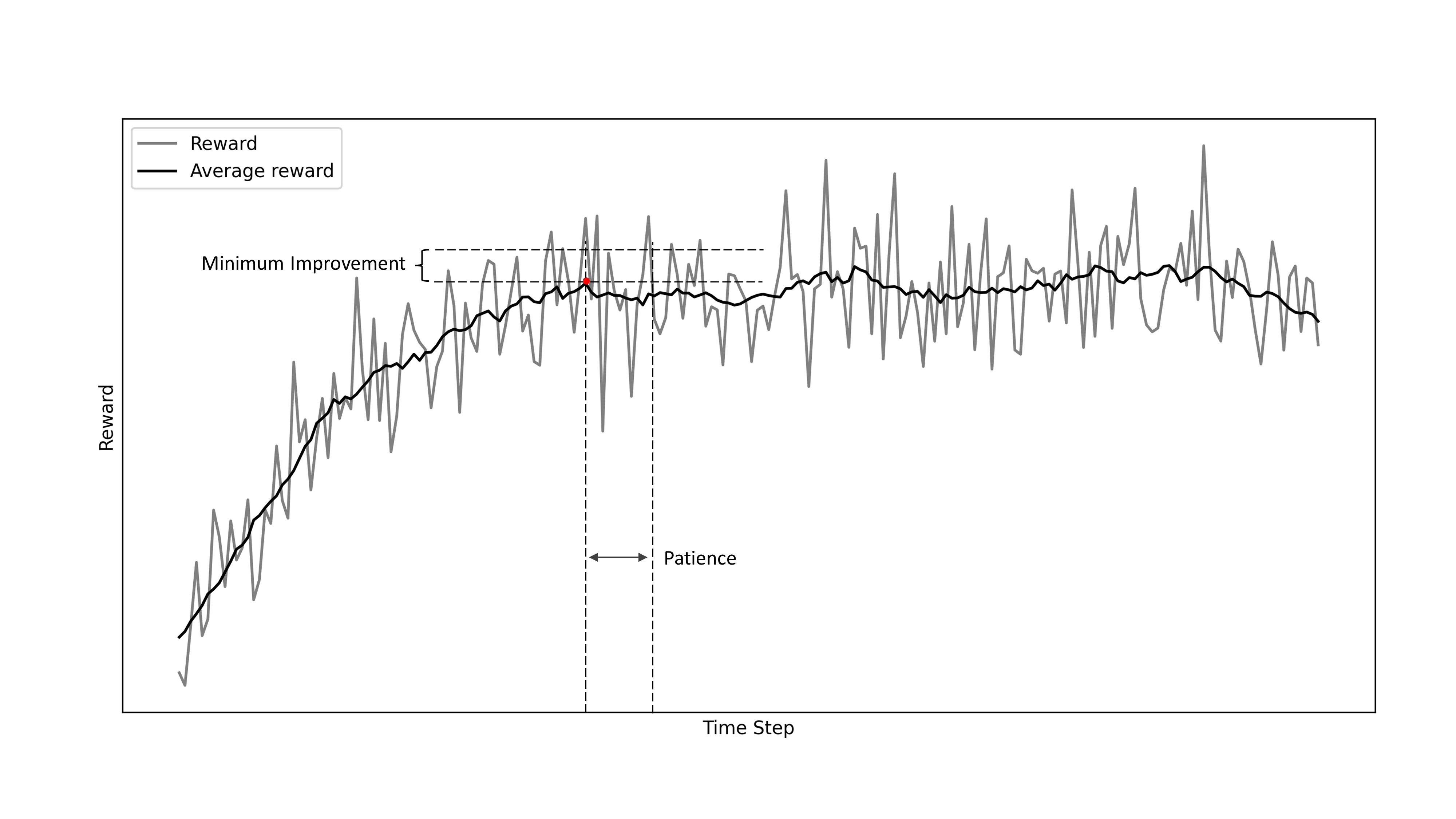}
    \caption{Average reward improvement rate based on early forecasting plot.}
    \label{fig:early_stopping}
    \vspace{-0.1cm}
\end{figure*}

One of the primary challenges in reinforcement learning lies in its training duration. As the model progresses in learning, its accuracy tends to improve at a slower pace, eventually reaching a plateau where further training may even degrade performance, a phenomenon commonly known as overfitting. Stopping the model training at the appropriate state can mitigate both prolonged training times and overfitting risks. Implementing early stopping in reinforcement learning encounters specific challenges, notably the increased noise inherent in reward signals, as depicted in Figure \ref{fig:early_stopping}. Additionally, as training progresses, the rate of improvement decreases, and such progress is not always guaranteed. To address these challenges, this work introduces a two-staged early stopping process. The averaged rewards are computed to smooth the improvement pattern in the first stage. The moving average smoothing equation is given by equation (\ref{eq:mov_avg}):
\begin{equation}
sr_t = \frac{1}{n} \sum_{i=0}^{n-1} r_{t-i} \label{eq:mov_avg}
\end{equation}
where \( sr_t \) represents the smoothed reward point at time step \( t \), \( n \) is the window size, and \( r_t \) are the original reward points within the window centered around time step \( t \). \\

The second part involves the implementation of the actual early stopping mechanism, which incorporates a patience counter. This counter is incremented each time the new average reward falls short of the best-achieved reward. The number of episodes to wait, known as the patience number, is optimized during the hyperparameter tuning phase. Whenever a new reward exceeds the previous best reward, the counter is reset to zero. However, in addition to merely resetting the counter when a new reward exceeds the previous best, the reward must exceed the best previous average reward by a certain minimum improvement value, known as the improvement rate (IR). This ensures that the model's performance exhibits a substantial improvement before continuing training, thereby enhancing efficiency and preventing premature termination. the improvement rate is given by equation (\ref{eq:improvement_rate}):

\begin{equation}
\text{IR} = \frac{\text{Current Average Reward}}{\text{Best Average Reward}}\label{eq:improvement_rate}
\end{equation}
This ensures that unless a minimum improvement difference within a specified number of episodes can be observed, the model should not continue training.

\section{Empirical Study}

\begin{figure}[htbp]
    \centering
    \includegraphics[width=0.5\textwidth, angle=0,trim={10cm 3cm 10cm 1cm},clip]{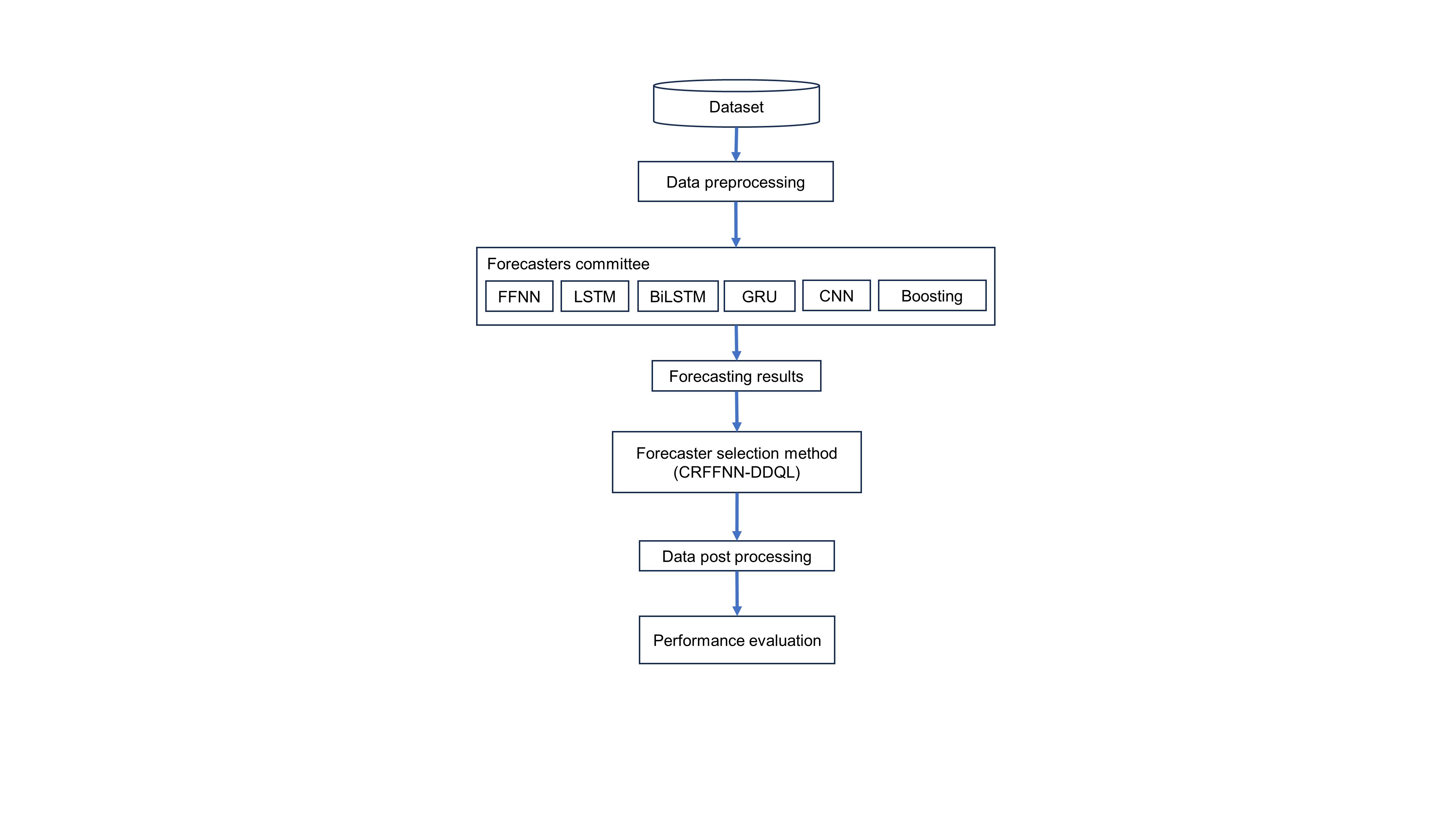}
    \caption{The experimental setup.}
    \label{fig:experimental_setup_fig_1_diagram}
    \vspace{-0.1cm}
\end{figure}

To assess the performance of the reinforcement learning-based forecast model selection, we developed the architectural configuration of experiments depicted in Figure \ref{fig:experimental_setup_fig_1_diagram}. Subsequent sections provide details of every step for this configuration.

\subsection{Datasets}

The study presented in the previous section of this article has been performed to design the best solution approach in selecting the appropriate forecasting method that predicts demand for problems of supply chain optimization. However, the datasets of demands for goods of a distribution company in France for whom we designed the solution approach presented in this article have been tagged as private. Therefore, one public dataset has been tested as well.

\subsubsection{Corporación favorita grocery sales dataset}

Derived from Kaggle's challenges focused on predicting future sales for a comprehensive range of fast-moving consumer goods, this study has narrowed its scope to the initial 200 products. These products are tracked from January 1st, 2013, to March 3rd, 2014. The forecasting task aims at predicting sales for the period spanning March 4th to March 31st, 2014, covering a 28-day horizon. It's important to note that all data values within this dataset are integers.

\subsubsection{Client snacks demands dataset}

The dataset, provided by a client, comprises over 10,000 time series of snack demand, each recorded daily and spanning 255 data points. For this study, the focus was narrowed to the initial 200 products, with a forecasting horizon of 28 days. Notably, all values within this dataset are integers.

\subsection{Data pre-processing}

Before feeding the data into our models, a series of pre-processing steps have been meticulously applied, leveraging the methodology outlined in a previous publication \cite{HEWAMALAGE2021388}. Each time series underwent an initial partitioning into training, validation, and testing subsets, resulting in three distinct datasets. Each model is trained using the train split and validated using the validation split. The hyperparameters tuning process does not run on the test split. Once the best model has been selected, this model is trained again on both the train and validation sets. The forecasting error is finally measured on the test data set. In both datasets, to fill in missing values, a straightforward substitution method was employed, replacing all missing values with zeros. \\

In contrast to previous research highlighting the importance of unseasonalization \cite{claveria_data_2017}, this study refrained from applying such techniques. This decision was informed by earlier findings, as it was observed that both datasets with minimal seasonality or sharing uniform seasonal patterns could be modeled by neural networks without unseasonalization \cite{HEWAMALAGE2021388}. Each time series was normalized by dividing values by their mean, which helps standardization of all time series in a common range, easing the learning process. Since the dataset feeding this study does not contain negative values, a straightforward logarithmic transformation was applied to stabilize the variance, as shown in equation (\ref{eq:equation_log}):

\begin{equation}
\text{w\textsubscript{t}} = \begin{cases}
\log(\text{y\textsubscript{t}} + 1) & \text{if } \min(y) \leq 0 \\
\log(\text{y\textsubscript{t}}) & \text{if } \min(y) > 0
\end{cases}
\label{eq:equation_log}
\end{equation}

\noindent In this context, the \textit{y} variable represents the whole sequence, and for this specific case, the value of \textit{ε} (epsilon) is set to zero since all data points are equal to or greater than zero. The \textit{c} constant is also set to 1 in this instance. To achieve multi-step-ahead forecasting, a recursive strategy, which is a sequence of one-step-ahead forecasts has been used. It involves feeding the prediction from the last time step as an input to the next prediction.

\subsection{Training and validation}

Each model follows a two-step procedure involving training and validation. The model configuration that demonstrates superior performance, meaning the lowest validation error, is subsequently selected, taking into account both the training and validation splits. This selected model is then evaluated on the independent test split to assess its generalization capability. Furthermore, the study presented in this article leverages the concept of global models, where one model is trained on the full dataset, enabling the capture of broader patterns and trends from data. The experiments whose results are presented below run on two Quadro RTX 8000 GPUs, each equipped with 51GB of memory.

\subsection{Post processing}

Following the forecasting process, post-processing is required to derive the actual values from the forecasted data. This step mainly stands for reversing the pre-processing procedures. Initially, the exponential of the predicted time series is calculated to revert the earlier applied log transformation. Additionally, the value \textit{1} is subtracted from each data point if the time series originally contains zero values. Each forecasted time series is then scaled by its mean (previously computed during pre-processing). Finally, each data point is rounded up to the closest integer, and any negative value is clipped to zero.

\subsection{Error metrics}

To assess the quality of the models, a comparison between the predicted values and the ground truth is required. For this purpose, each model is trained on both the training and validation data and then tested on the test data. This training and testing process is repeated ten times with different seeds. Subsequently, the errors for each iteration are computed using two distinct evaluation metrics:

\subsubsection{The Symmetric Mean Absolute Percentage Error (SMAPE)}

The SMAPE is widely used in forecasting models and time series applications in general. This metric provides results in percentage terms, which are easy to interpret and communicate to stakeholders. This makes it more meaningful than other error metrics that may not be expressed by intuitive units.
\begin{equation}
\text{SMAPE} = \frac{100\%}{n} \sum_{t=1}^{n} \frac{|A_t - F_t|}{(|A_t| + |F_t|)/2}
\end{equation}
\noindent Where F\textsubscript{t} is the forecast value, A\textsubscript{t} is the Actual value, and n is the number of samples (the forecast horizon). However, since both datasets we are using have values close to zero which makes the SMAPE results unstable, a variant of the SMAPE has been suggested by \cite{suilin_kaggle_2023} and applied by \cite{HEWAMALAGE2021388}. As these experiments follow the same experimental protocol, we monitor the same metric except that the denominator is replaced by the equation below:
\begin{equation}
\max \left( |A\textsubscript{t}| + |F\textsubscript{t}| + \epsilon, 0.5 + \epsilon \right)
\end{equation}
Where epsilon was set to $0.1$. This helps avoid division by values close to zero

\subsubsection{The Mean Absolute Scaled Error (MASE)}

\begin{equation}
\text{MASE} = \frac{1}{n} \sum_{t=1}^{n} \frac{|A\textsubscript{t} - F\textsubscript{t}|}{\frac{1}{n-1} \sum_{t=2}^{n} |A\textsubscript{t} - A\textsubscript{t-1}|}
\end{equation}
$\sum_{t=2}^{n}|A_{t} - A_{t-1}|$ calculates the sum of absolute differences between consecutive actual values, which represent the mean absolute difference between consecutive observations. It is applied as the denominator to scale the errors. The MASE error is then a scale-independent metric that is crucial when comparing forecasting models or evaluating forecast accuracy across different datasets with varying scales.

\subsection{Benchmarks}

For the benchmarks, The forecasters' committee has been selected to compare the suggested solution performance against a stand-alone model. This includes :

\begin{enumerate}
    \item The Feed Forward Neural Network (FFNN), the most prevalent type of artificial neural network in literature, is designed to tackle complex problems by simulating the workings of the human brain. At its early stage, the perceptron, first introduced by Rosenblatt \cite{rosenblatt_perceptron_1958}, serves as the fundamental unit of FFNN.
    \item Long Short-Term Memory (LSTM), a model specifically tailored for handling sequential data, has emerged as a powerful tool for various applications, including demand forecasting \cite{LSTM_paper}.
    \item The Gated Recurrent Unit (GRU) is closely related to LSTM but features a simpler architecture, resulting in faster training \cite{cho-etal-2014-learning}.
    \item Bidirectional LSTM (BiLSTM) processes input sequences in both forward and backward directions simultaneously, enhancing data fitting capabilities \cite{bilstm}.
    \item The Convolutional Neural Network (CNN) was initially introduced in 1998 for recognizing handwritten digits \cite{lecun_gradient-based_1998}. Its capability to capture spatial features makes it an appealing choice for forecasting tasks.
    \item Boosting, in the field of machine learning, is an ensemble meta-algorithm primarily designed to reduce bias and, concurrently, address variance in supervised learning. It falls within a family of machine learning algorithms dedicated to enhancing the performance of weak learners, allowing models to compensate for each other's weaknesses \cite{bishop_pattern_2006}. Recognizing LSTM's proficiency in capturing temporal features and CNN's strength in handling spatial features, combining both could lead to better modeling of both types of features.
\end{enumerate}

To compare against multiple forecasters selection strategies, this work has also been compared against multiple strategies of model selection:

\begin{enumerate}
    \item Ensemble averaging (Mean): This method simply computes the mean of multiple models to have a more stable result. This helps reduce the bias \cite{perrone_when_1995}.
    \item Median ensembling, also known as Median, is a closely related approach to the mean. However, rather than computing the mean of the models' outputs, it computes the median.
    \item Stacking: often referred to as "Stacked Generalization," is a potent ensemble machine learning technique intended to boost a model's overall performance \cite{wolpert_stacked_1992}. At its core, stacking involves the integration of multiple base models, with each base model generating individual predictions. A higher-level model uses as input these individual predictions, known as the meta-model or blender. The term "meta-model" denotes a higher-level model trained to make predictions or decisions based on the outputs of multiple base models, strategically combining them to produce the final forecast. This ensemble approach harnesses the diverse strengths of the base models, enabling more robust and accurate predictions.
    \item Double Deep Q-learning (DDQL) shares similarities with the stacking technique, but instead of applying a supervised learning approach, reinforcement learning runs. In this framework, an agent learns in order to maximize its reward by selecting the best forecaster from the forecasters' committee at a given time t \cite{ddql_paper}.

\end{enumerate}

\section{Results and discussion}

This section presents a thorough analysis of the experimental outcomes on the two sales datasets already mentioned (Corporación Favorita Grocery Sales and client snacks demands). Subsequent sections delve into the specific results derived from each dataset.

\subsection{The results from the Corporación favorita dataset}

To ensure consistency of the results, each experiment ran 10 times. The experimental results of the corporación favorita grocery sales dataset are recorded in Table~\ref{tab:model_comparison_favorita} where the mean of each model is computed from 10 executions. The well-known methods. The comparison has been carried out against six neural network forecasting methods and four model selection methods. Computing the forecast error on 28 days using a sequence of one-step-ahead forecasts.

\begin{table}[htbp]
\centering
\caption{Favorita Dataset Mean MASE and Mean SMAPE errors.}
\label{tab:model_comparison_favorita}
\scriptsize
\begin{tabular*}{\linewidth}{@{\extracolsep{\fill}}p{3.2cm}p{1.4cm}p{1.4cm}p{1.4cm}p{1.4cm}@{}}
\toprule
\textbf{Model} & \textbf{Mean SMAPE} & \textbf{Std. Div. SMAPE} & \textbf{Mean MASE} & \textbf{Std. Div. MASE} \\ \midrule
FFNN & 37.81 & 10.21 & 1.84 & 0.25 \\
LSTM & 38.99 & 10.23 & 1.92 & 0.31 \\
GRU & 38.46 & 10.12 & 1.87 & 0.25 \\
BiLSTM & 38.93 & 10.16 & 1.90 & 0.24 \\
CNN & 38.19 & 10.44 & 1.85 & 0.31 \\
CNN-LSTM & 39.07 & 10.47 & 1.91 & 0.32 \\ \midrule
Mean & 40.92 & 11.20 & 1.97 & 0.36 \\
Median & 38.32 & 10.18 & 1.86 & 0.26 \\
Stacking & 37.77 & 9.06 & 1.83 & 0.23 \\
FFNN-DDQL & 37.59 & 10.01 & 1.89 & 0.25 \\
\textbf{CRFFNN-ARIRBES} & \textbf{37.03} & \textbf{4.58} & \textbf{1.79} & \textbf{0.17} \\ \bottomrule
\end{tabular*}
\end{table}

Among the six forecasters in the committee, the FFNN model discloses a lower forecasting error on both SMAPE and MASE metrics compared to the other models. However, about stability, the standard deviation of forecasting errors for the GRU model is lower on the SMAPE metric, while the BiLSTM model demonstrates greater stability on the MASE metric. A lower standard deviation suggests a more stable convergence for the model. Notably, among the forecast model selection methods, stacking, FFNN-DDQL, and the proposed CRFFNN-ARIRBES method achieves lower forecasting errors on both the SMAPE and MASE metrics. Regarding standard deviation, the suggested CRFFNN-ARIRBES method yields the best results, with the stacking method ranking second. These findings underscore the robustness of the suggested architecture. \\

\begin{figure*}[htbp]
    \centering
    \includegraphics[width=0.9\textwidth, angle=0, trim={0cm 0cm 0cm 0cm}, clip]{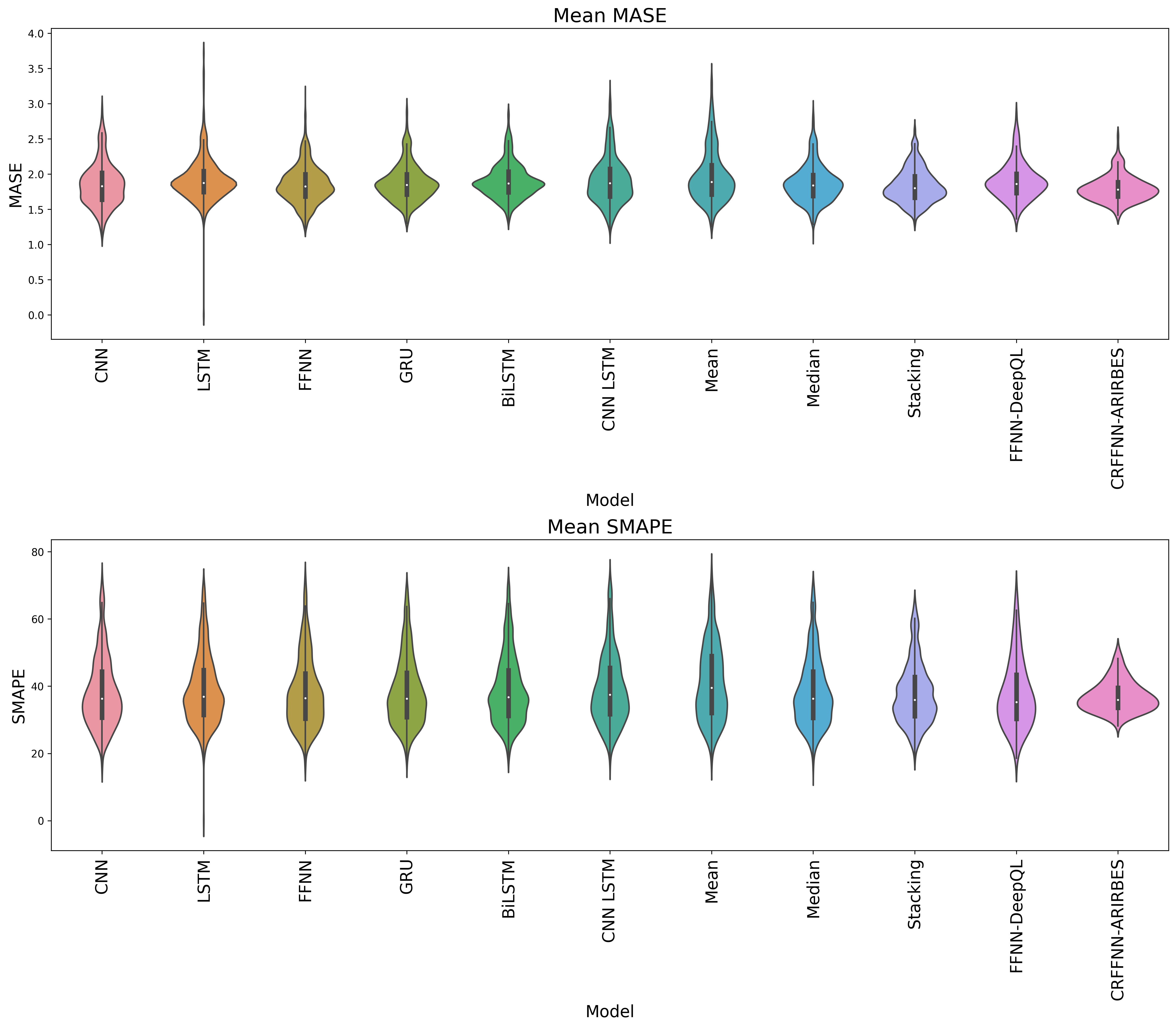}
    \caption{MASE and SMAPE Errors for NN5 Dataset}
    \label{fig:results_favorita_violin}
    \vspace{-0.1cm}
\end{figure*}

Leveraging the violin plot, a visually immersive tool encapsulating data distribution intricacies, we glean insights into central tendency, spread, and probability density. This dual-functionality plot emerges as a valuable instrument for data comparison and interpretation in research. The violin plots showcased in Fig. \ref{fig:results_favorita_violin} disclose a comprehensive view of the performance distribution of each model's errors across different random weight initializations. This visual representation is instrumental in discerning the stability nuances inherent to the models. By scrutinizing the spread of errors, we can glean valuable insights into how robustly each model converges to its lowest achievable error under various initialization scenarios. \\

A noteworthy observation is centered around the Stacking, FFNN-DDQL, and the suggested CRFFNN-ARIRBES methods. The violin plots for these three show a remarkably narrow spread of errors, accompanied by a graph predominantly situated at the lower end. This convergence pattern signifies a high degree of stability, suggesting that these methods consistently converge to a low error irrespective of the specific initial conditions. Out of these three methods, the violin plot of the suggested CRFFNN-ARIRBES method showcases a slightly slender error distribution compared to stacking and FFNN-DDQL. This consistency in achieving low errors points out the capability of the suggested CRFFNN  architecture used as the DDQL agent at capturing the spatial and temporal features from the environment. \\

In summary, the nuanced analysis of the violin plots provides a deeper understanding of each model's performance and stability. The observed convergence patterns show valuable insights for model selection and highlight potential areas for improvement, contributing to the robustness and reliability of machine learning applications.

To validate the usefulness of the suggested early stopping methods ARIBES, The same suggested CRFFNN method is implemented with early stopping CRFFNN-ARIBES. The training time and forecast errors are summarized in Table \ref{tab:early_stoppinh_table_favorita}
\begin{table}[htbp]
    \centering
    \caption{Favorita Dataset training time and forecast errors.}
    \label{tab:early_stoppinh_table_favorita}
    \scriptsize
    \begin{tabular*}{\linewidth}{@{\extracolsep{\fill}}p{3.2cm}p{1.4cm}p{1.4cm}p{1.4cm}p{1.4cm}p{1.4cm}@{}}
        \toprule
        Model & Training Time (s) & Mean SMAPE & Std. Div. SMAPE & Mean MASE & Std. Div. MASE \\
        \midrule
        CRFFNN & 19841 & 36.97 & 4.37 & 1.83 & 0.16 \\
        CRFFNN-ARIBES & 5120 & 37.03 & 4.58 & 1.79 & 0.17 \\
        \bottomrule
    \end{tabular*}
\end{table}

The results from table ~\ref{tab:early_stoppinh_table_favorita} demonstrate the efficiency of the ARIRBES in reducing the training time. This lead to considerably cutting down the training cost while keeping a lower forecast error than the other methods.

\subsection{The results from the client snacks demands dataset}

The same experiment was conducted on another dataset offered by a client. This dataset has daily demands for snacks. The results of this experiment are documented in Table~\ref{tab:model_comparison_casestudy}.

\begin{table}[htbp]
\centering
\caption{Case Study Mean MASE and Mean SMAPE forecast errors.}
\label{tab:model_comparison_casestudy}
\scriptsize
\begin{tabular*}{\linewidth}{@{\extracolsep{\fill}}p{3.2cm}p{1.4cm}p{1.4cm}p{1.4cm}p{1.4cm}@{}}
\toprule
\textbf{Model} & \textbf{Mean SMAPE} & \textbf{Std. Div. SMAPE} & \textbf{Mean MASE} & \textbf{Std. Div. MASE} \\ \midrule
FFNN & 56.15 & 12.61 & 1.60 & 0.46 \\
LSTM & 65.81 & 15.36 & 2.09 & 0.48 \\
GRU & 64.25 & 15.38 & 1.80 & 0.49 \\
BiLSTM & 63.97 & 14.34 & 1.91 & 0.45 \\
CNN & 49.36 & 19.07 & 1.59 & 0.51 \\
CNN-LSTM & 65.66 & 16.36 & 1.95 & 0.55 \\ \midrule
Mean & 62.66 & 15.05 & 1.78 & 0.47 \\
Median & 63.30 & 15.26 & 1.80 & 0.47 \\
Stacking & 49.30 & 10.66 & 1.57 & 0.40 \\
FFNN-DDQL & 49.35 & 9.75 & 1.55 & 0.33 \\
\textbf{CRFFNN-ARIRBES} & \textbf{48.52} & \textbf{7.64} & \textbf{1.49} & \textbf{0.25} \\ \bottomrule
\end{tabular*}
\end{table}

The results from Table~\ref{tab:model_comparison_casestudy} highlight the capability of the CNN model to achieve lower forecasting error compared to the other forecasters on both the SMAPE and MASE metrics. In terms of stability, the FFNN achieves the best stability by having a lower standard deviation for both the MASE and SMAPE errors. Out of the forecasting selection methods, the stacking, FFNN-DDQL, and the suggested CRFFNN-ARIRBES method achieved lower forecasting errors on both the SMAPE and MASE metrics. In terms of standard deviation, the suggested CRFFNN-ARIRBES method achieved the best results, the FFNN-DDQL method is ranked second. This showcases the robustness of the suggested CRFFNN-ARIRBES architecture. \\

\begin{figure*}[htbp]
    \centering
    \includegraphics[width=0.9\textwidth, angle=0, trim={0cm 0cm 0cm 0cm}, clip]{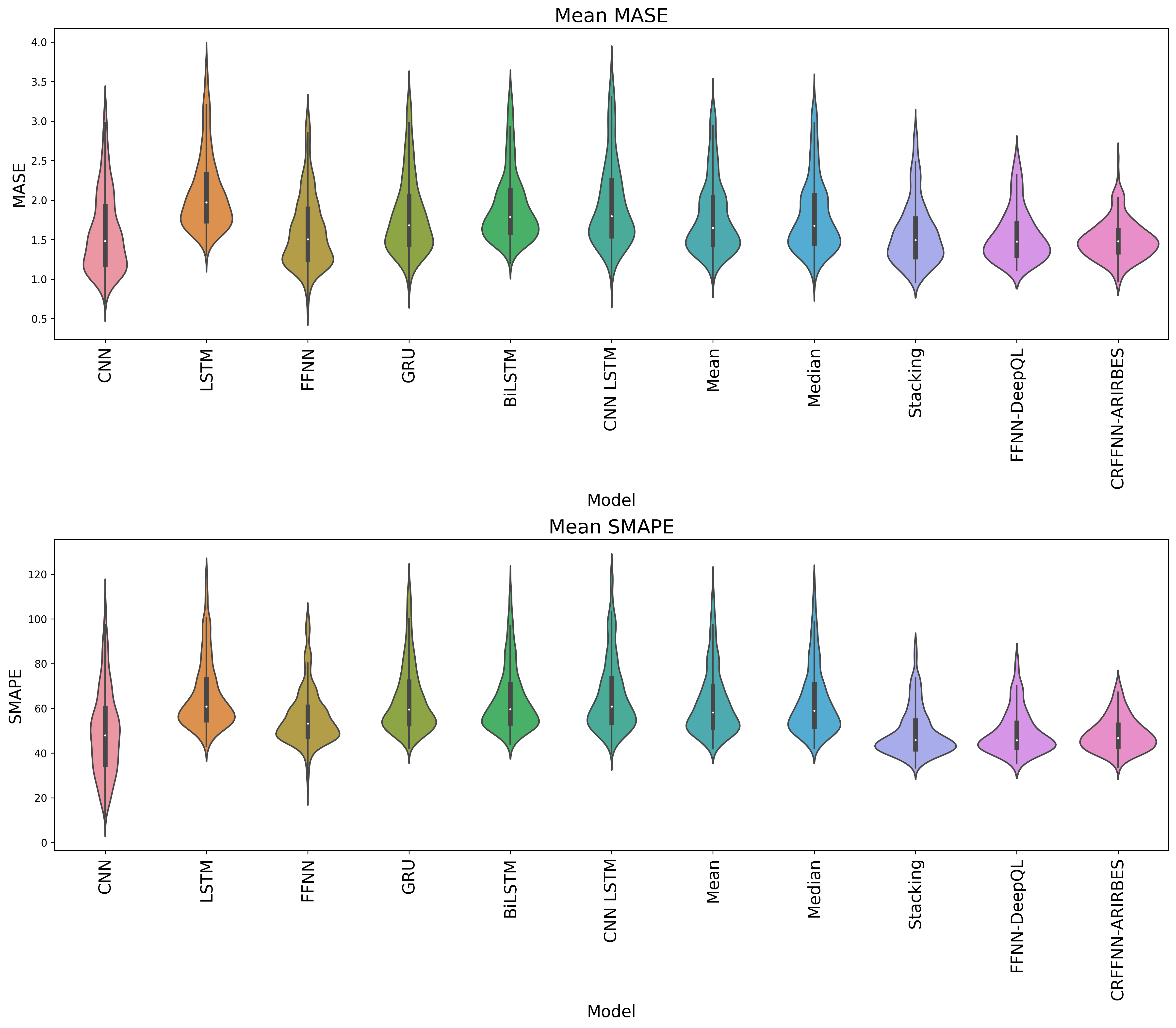}
    \caption{MASE and SMAPE Errors for Case Study Dataset}
    \label{fig:results_casestudy_violin}
    \vspace{-0.1cm}
\end{figure*}

The analysis of the violin plots in Fig. \ref{fig:results_casestudy_violin}, depicting the 10 errors for each model, offers valuable insights into their stability. The diverse random weight initialization provides a lens through which we can assess each model's convergence towards the lowest possible error. \\

Observing the violin plots of the 10 errors in Fig. \ref{fig:results_casestudy_violin} for each model provides valuable insights into the stability of each model. Running each model with different initial random weights helps grasp its ability to converge towards the lowest possible error. In this context, it's worth highlighting that the stacking, FFNN-DDQL, and the suggested CRFFNN-ARIRBES method disclose higher convergence to lower errors compared to the forecasters and the mean and median selection method. this highlights the robustness of these three methods. Out of these three methods, the suggested CRFFNN-ARIRBES solution stands out, achieving the lowest error and displaying minimal error disparity in the violin plots. This underscores its high stability in convergence, making it a compelling choice for applications requiring consistently low errors across diverse initialization conditions. The same early stopping strategy ARIRBES has been applied on the client dataset as well. The training time and forecast errors are resumed in the table  ~\ref{tab:training_time_casestudy}. \\

\begin{table}[htbp]
    \centering
    \caption{Case study Dataset training time.}
    \label{tab:training_time_casestudy}
    \scriptsize
    \begin{tabular*}{\linewidth}{@{\extracolsep{\fill}}p{3.2cm}p{1.4cm}p{1.4cm}p{1.4cm}p{1.4cm}p{1.4cm}@{}}
        \toprule
        Model & Training Time (s) & Mean SMAPE & Std. Div. SMAPE & Mean MASE & Std. Div. MASE \\
        \midrule
        CRFFNN & 10190 & 48.74 & 7.82 & 1.46 & 0.23 \\
        CRFFNN-ARIBES & 2635 & 48.52 & 7.64 & 1.49 & 0.25 \\

        \bottomrule
    \end{tabular*}
\end{table}

The results from table ~\ref{tab:training_time_casestudy} show the efficiency of the ARIRBES in reducing the training time. This lead to considerably cutting down the training cost while still reducing the prediction error in comparison to the other models. \\

From the analysis of both datasets, it is obvious that each model's performance can vary based on the dataset or initial weights. This underscores the effectiveness of reinforcement learning in reducing error by selecting the model with the lowest error at any given time. A notable limitation of the current approach lies in the training and inference time required. Considering that each model within the forecasters' committee must run during both the training and inference phases, the process becomes computationally intensive. Furthermore, an major factor in enhancing forecast quality is to increase the dataset size, which inevitably involves a significant impact on the cost of forecasting solutions.

\section{Conclusion and discussions}
In supply chain management, forecasting demand can be considered as the first step of a sequential process that impacts multiple subsequent decisions. Stability and accuracy of the forecasted results have become an important challenge for many distribution and transportation companies whose profits can directly depend on the fluidity of their processes. \\

In the field of demand forecasting, a wide array of diverse datasets exists, each one having its own features such as seasonality or noise levels. The performance of forecasting models varies across these datasets, making the selection of the most suitable model for a given time series a complex task, often relying on manual trial-and-error by experts in the forecasting domain. To address this challenge automatic model selection techniques have been developed over the years. These techniques predominantly rely on supervised learning methods such as linear regression and support vector machines to select the optimal forecasting model at each step. However, these approaches have inherent limitations in their ability to adapt to evolving data patterns over time. In response to this limitation, reinforcement learning techniques, such as Q-learning for continuous model selection, have been introduced and tested for a few years. \\

To leverage the adaptability of reinforcement learning in accommodating changes in forecasting patterns and the deep learning capacity for modeling non-discrete values, a innovating deep neural architecture has been developed in this study. This architecture is based on convolution, recurrent, and feed-forward neural networks as a double deep reinforcement learning agent  This enables better spatial and temporal feature modeling. Another significant contribution from this article deals with an average reward-based early-stopping approach that has been computed to prevent the model from getting stuck during extended training. This selection model of forecasting has been built and tested to perform on two databases of grocery sales. The first one is public and already used to assess performances of other new forecasting models and the second one comes from actual data from a company delivering software to support transportation companies in the grocery field. \\

Experimental results show that the suggested method outperforms all the forecasting methods and selection methods on both datasets table \ref{tab:model_comparison_favorita} and \ref{tab:model_comparison_casestudy}, with regards to Mean SMAPE and Mean MASE criteria. From experiments, we observe on the Favorita instance that the Mean SMAPE (resp. Mean MASE) computed from the suggested forecasting approach is almost $37.03$ (resp. $1.79$), although the values computed by other approaches vary between $37.59$ and $40.92$ (resp. $1.83$ and $1.97$). The standard deviation of the SMAPE (resp. MASE) computed from the suggested forecasting approach is almost $4.58$ (resp. $0.17$), although the values from other approaches vary between $9.06$ and $11.20$ (resp. $0.23$ and $0.36$). From the client snacks dataset, results are also favorable to the suggested approach. The Mean SMAPE (resp. Mean MASE) reaches $48.52$ (resp. $1.49$) although others take values between $49.30$ and $56.15$ (resp. $1.55$ and $2.09$). The standard deviation of the SMAPE (resp. MASE) computed from the suggested forecasting approach is almost $7.64$ (resp. $0.25$), although the values from other approaches vary between $9.75$ and $19.07$ (resp. $0.33$ and $0.55$). The graphical displays in Fig.\ref{fig:results_casestudy_violin} and Fig.\ref{fig:results_favorita_violin} further highlight smaller errors after each model run and evaluations, meaning improved convergence of the deep learning phase towards the optimal forecasting selection model. These figures also show, looking at the violin shapes, that the CRFFNN model is more stable than other approaches when initial inputs vary. In summary, the analysis of experimental results tends to prove the viability and efficiency of automatic model selection techniques in the context of demand forecasting. The effectiveness of the early stopping approach ARIBES has been tested as well. Adding this stopping criterion reduces the training time from 19841s to 5129s for the favorita dataset and from 10190s to 2635s for the snacks demand dataset. \\

Following the analysis performed and detailed in this articles, potential way for future research arises. A first one is about enhancement of forecasting by including additional explanatory variables into the model. Some architecture adjustment would then be required. We are exploring the addition of macroeconomic indicators into the set of endogenous variables, particularly when they are relevant to specific groups of products. Additionally, investigating the benefits of segmenting datasets into smaller, more homogeneous subgroups could be worth it. This approach might help alleviate forecast errors while still maintaining an efficient training time. Another way of improving the proposed approach concerns the reduction of the training cost, especially when this approach is applied to large datasets. Training the whole forecasters' committee, coupled with training in the reinforcement learning-based selection method, consumes a significant amount of computer resources and may significantly increase business expenses. Future endeavors should prioritize addressing this challenge to make the approach more feasible for practical applications.

\end{document}